\documentclass[a4paper,12pt]{article}
\usepackage{a4wide}
\usepackage{times}
\usepackage[utf8]{inputenc}
\usepackage[T1]{fontenc}
\usepackage{eurosym}
\usepackage[usenames,dvipsnames]{color}
\usepackage{graphicx}
\usepackage{import}

\usepackage{setspace}
\usepackage{wrapfig}

\usepackage{stfloats} 		
\usepackage{fancyhdr}
\usepackage{hyphenat}		
\usepackage{url}
\usepackage{fncylab}
\usepackage{xspace}
\usepackage{icomma}
\usepackage[labelfont={sf,bf},textfont={sf},font=small]{caption}

\usepackage{tabularx}
\usepackage{enumitem}
\setlist{nolistsep}

\usepackage[english]{babel}

\usepackage{multirow}

\usepackage[bookmarks,				
            bookmarksopen=false,	
            bookmarksnumbered=true,	
            pdfstartview=Fit, 		
            linkbordercolor={0 0 1},		
            citebordercolor={0 0.80 0.2},	
            urlbordercolor={0.55 0.11 0.38},	
            colorlinks,			
            linkcolor=Black,	
            citecolor=Blue,		
            urlcolor=Blue,		
			      filecolor=Blue 		
			      plainpages=false,
			      pdfpagelabels]{hyperref}	

\usepackage{amsmath}

\graphicspath{{./Grafiken/}{./}}
\DeclareGraphicsExtensions{.pdf}   

\newenvironment{arbp}{%
\medskip\noindent\tabularx{\textwidth}{@{\hspace{12pt}} l X@{}}%
}{%
\endtabularx%
}

\begin{document}
\title{	\begin{normalsize}
		\begin{flushleft}
			********************\\
			This documentation of the industrial benchmark (IB) is \color{red}\textbf{deprecated}\color{black}. 
			The current documentation can be found at \url{https://arxiv.org/abs/1709.09480}.\\
			********************\\
			\medskip
			\medskip
		\end{flushleft}
		\hrulefill
		\color{white}
		.\\
	\end{normalsize}
	\color{black}
	Introduction to the ``Industrial Benchmark''}
\author{Daniel Hein, Alexander Hentschel, Volkmar Sterzing, Michel Tokic,\\
Steffen Udluft
\vspace{.3cm}\\
%
Siemens AG, CT RTC BAM LSY-DE\\
Learning Systems
}
\maketitle

\begin{abstract}
A novel reinforcement learning benchmark, called Industrial Benchmark, is introduced.
The Industrial Benchmark aims at being be realistic in the sense, 
that it includes a variety of aspects that we found to be vital in industrial applications.
It is not designed to be an approximation of any real system, but to pose the same hardness and complexity.
\end{abstract}

\section{Introduction}
The scientific method requires that a hypothesis is tested by experiments.
This holds true in the field of machine learning, when algorithms are to be
developed or improved. 
Such a test can be, to run the algorithm on a real system, in order to observe its performance,
or to run it on a simulation, a virtual system implemented as a computer program.
The latter method has several advantages: 
it is usually faster, cheaper, and of course more safe
to test the algorithm on a simulation. In addition the simulation can be manipulated much more freely than a real system.
Internal states can be observed, stored, restored, and set, 
thus given the freedom to test special aspects with little effort.
The amount of data can usually be enlarged such that all results gain statistical significance.
This stands in drastic contrast to the situation when testing the algorithm on a large scale industrial system, 
like a power plant.

There are some disadvantages with simulation based testing though.
First of all, the final success of an improved algorithm will be defined by the performance on the real system.
Thus any deviation of the simulation from the real system might cause the development of new algorithms and the 
process of improving to go not in the right direction.
The simulation might be too simple, underestimating the challenges posed by the real system.
Or, it might focus on less relevant aspects of the task, posing artificial difficulties.
Therefore it would be desirable to use a most realistic simulation for benchmarking.
On the other hand, when developing an algorithm one usually targets for methods that 
are applicable for a broad spectrum of systems and fine tuning for a specific system 
is fruitful from the perspective of that system only, 
while it does not help to decide for the algorithm which is ``superior in general''.
It can be doubted that it is possible to create an algorithm that is ``superior in general''.
Usually special cases exist where the seemingly inferior method is superior, 
but still it seems good practice to aim for a method that performs well in a wide variety of simulation 
benchmarks.
This dilemma of specific versus general solutions will not be solved here, 
but we want to contribute 
a software benchmark\footnote{Java source code: \url{http://github.com/siemens/industrialbenchmark}} 
that captures many aspects that we found to be vital in 
industrial applications.
The basic task, we are considering is the optimization of operating an industrial system.
This task will be described in the theoretical framework of reinforcement learning \cite{sutton_and_barto1998rl}.
The proposed simulation, called Industrial Benchmark, will serve as environment, i.e. the system to be controlled.
Independently of reinforcement learning, the Industrial Benchmark can also be used to evaluate regression, forecasting, and system identification
capabilities of different machine learning methods, as well as specific challenges like transfer learning, 
active learning, feature selection, or change detection.



\section{Industrial Benchmark}
The Industrial Benchmark aims at being realistic in the sense, that it includes a variety of aspects that 
we found to be vital in industrial applications.
It is not designed to be an approximation of any real system, but to pose the same hardness and complexity.
State- and action-space are continuous, the state-space is rather high-dimensional, and only partially observable.
The actions consist of three continuous components and effect three steerings. 
There are delayed effects. 
The optimization task is multi-criterial in the sense that there are two reward-components 
that show opposite dependencies on the actions. The dynamical behaviour is heteroscedastic with state dependent observation noise 
and state dependent probability distributions, based on latent variables. 
The dynamical behaviour is dependent on an external driver,
that cannot be influenced by the actions. 
The Industrial Benchmark is designed such that the optimal policy will not approach a fixed operation point in the three steerings.
Any specific choice is driven by our experience with industrial challenges.

\section{Detailed description}
At any time step $t$ the reinforcement learning agent can influence the environment (Industrial Benchmark) via actions $\vec{a}(t)$ 
that are three dimensional vectors in $[-1,1]^3$.
Each action can be interpreted as three proposed changes to the three observable state variables called current steerings.
Those current steerings are named velocity $v$, gain $g$, and shift $s$. Each of those is limited to $[0,100]$.
\begin{eqnarray}
\vec{a}(t) & = & (\Delta v,\Delta g, \Delta s)^{\top}\,,\\
\nonumber\\
v(t+1) & = & {\rm min}(0.{\rm max}(100,v + \Delta v))\,,\\
g(t+1) & = & {\rm min}(0.{\rm max}(100,g + 10 \, \Delta g))\,,\\
s(t+1) & = & {\rm min}(0.{\rm max}(100,s + \alpha_s \Delta s))\,.
\end{eqnarray}
Where the step size for changing shift is $\alpha_s = 20 \sin(15^0)/0.9\approx 5.75$.
After applying the action $\vec{a}(t)$ the environment transitions to the next time step $t+1$ 
in which it is in an internal state $\vec{s}(t+1)$\footnote{Not to be mixed up with the shift $s$, 
which is one of the variables that build the state $\vec{s}$.}.
State $\vec{s}(t)$ and successor state $\vec{s}(t+1)$ (also written as $\vec{s'}(t)$) are the markovian states of the environment that are only partially observable to the agent.

Three observable state variables we have already discussed, the current steerings $v$, gain $g$, and shift $s$.
There are three more observable state variables.
One is the external driver named set point $p$, that influences the dynamical behaviour 
of the environment but cannot be changed by the actions.
In the setting discussed here, the set point is kept constant (Constant Set Point Setting).
An extension will be that the set point will change with time (Variable Set Point Setting), i.e. $p=p(t)$.
In this Variable Set Point Setting the changes in the set point will be influenced externally and not by the actions.
In the Constant Set Point Setting the agent does not need to predict set point changes, as they do not occur.
This setting is therefore not one partially observable markov decision problem (POMDP), 
but a family (german: Schar) of POMDPs, parameterized by the set point. 
Learning to act optimal for any set point in the Constant Set Point Setting
can also be seen as a multi-task or transfer learning task \cite{Thrun96a,PanYang2010}.

The set of observable state variables is completed by the two reward relevant variables consumption $c(t)$ and fatigue $f(t)$.
In the general reinforcement learning setting a reward $r(t)$ is drawn for each transition $t\rightarrow t+1$ 
from state $\vec{s}(t)$ via action $\vec{a}(t)$ 
to the successor state $\vec{s}(t+1)$ from a probability distribution depending on $\vec{s}(t)$, $\vec{a}(t)$, 
and $\vec{S}(t+1)$.
In the Industrial Benchmark the reward is given by a deterministic function of the successor state $r(t)=r(\vec{s}(t+1))$, 
i.e.
\begin{equation}
\label{eq:reward}
r(t) = -(c(t+1) + f(t+1))\,.
\end{equation}
In the real world tasks that motivated the Industrial Benchmark the reward function has 
always been known explicitly.
In some cases it was subject to optimization itself and had to be adjusted to 
properly express the optimization goal.
For the Industrial Benchmark we therefore assume that the reward function is 
known and all variables influencing it are observable.

Thus the observation vector $\vec{O}(t)$ at time $t$ comprises current values of the set of observable state variables,
which is a subset of all the variables of state $\vec{s}(t)$, i.e.
\begin{enumerate}
\item the current steerings, velocity $v(t)$, gain $g(t)$, and shift $s(t)$,
\item the external driver, set point $p$,
\item and the reward relevant variables consumption $c(t)$ and fatigue $f(t)$.
\end{enumerate}

The data base for learning comprises of tuples $(\vec{O}(t),\vec{a}(t),\vec{O}(t+1),r(t))$, 
which, by introducing the notation $\vec{O}'$ for the observation vector of the successor state, 
will be written as $(\vec{O}(i),\vec{a}(i),\vec{O}'(i),r(i))$ or, in short, $(\vec{O},\vec{a},\vec{O}',r)(i)$.

The agent is allowed to use all previous observation vectors and actions to 
estimate the markovian state $\vec{s}(t)$.

\section{Description of the dynamical behaviour}
The dynamical behaviour of the Industrial Benchmark is determined by the three steerings velocity $v$, gain $g$, 
and shift $s$, 
the external driver set point $p$, and five latent variables.
The dynamics can be decomposed in three different sub-dynamics named operational costs, mis-calibration, and fatigue.
\subsection{Dynamics of operational cost}
The sub-dynamics of operational cost is influenced by the external driver set point $p$ and two of the three steerings, 
namely velocity $v$ and gain $g$. 
The current operational cost $o(t)$ is calculated as
\begin{equation}
o(t) = \exp\left(\frac{2 p(t) + 4 v(t) + 2.5 g(t)}{100}\right)\,.
\end{equation}
The current operational cost $o(t)$ cannot be observed, the observation is delayed and smeared out by a convolution
\begin{eqnarray}
o^c(t) & = & 0 o(t) + 0 o(t-1) + 0 o(t-2) + 0 o(t-3) + 0 o(t-4) + \nonumber\\
& &          \frac{1}{9} o(t-5) + \frac{2}{9} o(t-6) + \frac{3}{9} o(t-7) + \frac{2}{9} o(t-8) + \frac{1}{9} o(t-9) 
\end{eqnarray}
The convoluted operational cost $o^c(t)$ still cannot be observed directly, 
it is modified by the second sub-dynamics, called mis-calibration, and finally subject to observation noise.
The motivation for this dynamical bahaviour is that it is non-linear, depends on more than one influence, 
is delayed and smeared.
All those effects have been observed in industrial applications.

\subsection{Dynamics of mis-calibration}
\label{sec:Dynamics-of-mis-calibration}
The sub-dynamics of mis-calibration is influenced by the external driver, set point $p$, 
and only one steering, 
namely shift $s$.
Set point $p$ and shift $s$ are combined to an effective shift $s^e$
\begin{equation}
s^e = \min(1.5,\max(-1.5,s/20 - p/50 -1.5))\,,
\end{equation}
which influences the three latent variables $m_1^l$, $m_2^l$, and $m_3^l$.
The resulting mis-calibration $m$ is a function of effective shift and the latent variables
\begin{equation}
m = {\rm f}(s^e,m_1^l, m_2^l, m_3^l)\,.
\end{equation}


%
%
%

The resulting mis-calibration $m(t)$ is added to the convoluted operational cost $o^c(t)$, 
giving $\hat{c}$,
\begin{equation}
\hat{c} = o^c(t) + 25 m(t)\,,
\end{equation}
Before being observable as consumption $c$ the modified operational cost $\hat{c}$ 
is subject to heteroskedastic observation noise
\begin{equation}
c = \hat{c} + \rm{gauss}(0,1+0.02\,\hat{c})\,,
\end{equation}
i.e. a gaussian noise with zero mean and a standard deviation of $\sigma=1+0.02\,\hat{c}$.

\subsection{Dynamics of fatigue}
The sub-dynamic of fatigue is influenced by the same variables as the sub-dynamic of operational cost,
i.e. set point $p$, velocity $v$, and gain $g$. 
The Industrial Benchmark is designed in such a way that when changing the steerings velocity $v$ and gain $g$
as to reduce the operational cost, fatigue will be increased, leading to the desired
multi-criterial task, with two reward-components showing opposite dependencies on the actions.
The basic fatigue $f_b$ is computed as 
\begin{equation}
f_b = \max\left(0,\frac{30000}{5\,v + 100} - 0.01\, g^2\right)\,.
\end{equation}
From the basic fatigue $f_b$, the fatigue $f$ is calculated by
\begin{equation}
f = f_b (1 + 2 \alpha)\,,
\end{equation}
where $\alpha$ is an amplification.
The amplification depends on two latent variables $h_v$ and $h_g$,
an effective velocity $v^e$, an effective gain $g^e$, and is affected by noise,
\begin{equation}
  \alpha =
  \begin{cases}
    \frac{1}{1+\exp(-{\rm gauss(2.4,0.4)}} & \text{for}\quad \max(h_v,h_g)=1.2\\
    \max(\eta^v,\eta^g)       & \text{else}\,.
  \end{cases}
\end{equation}

The noise components $\eta^v$ and $\eta^g$,
as well as the latent variables $h_v$ and $h_g$,
depend on effective velocity $v^e$, and effective gain $g^e$.
These are calculated by set point dependent transformation functions
\begin{eqnarray}
{\rm T}_v(v,g,p) & = & \frac{g + p + 2}{v - p + 101}\,,\\
{\rm T}_g(g,p)   & = & \frac{1}{g + p + 1}\,.
\end{eqnarray}

Based on this transformation functions, effective velocity $v^e$ and effective gain $g^e$ are computet as follows:
\begin{eqnarray}
v^e & = & \frac{ {\rm T}_v(v,g,p) - {\rm T}_v(0,100,p)}
               { {\rm T}_v(100,0,p) - {\rm T}_v(0,100,p)}\\
g^e & = & \frac{ {\rm T}_g(g,p) - {\rm T}_g(100,p)}
               { {\rm T}_g(0,p) - {\rm T}_g(100,p)}\,.
\end{eqnarray}
To compute the noise components $\eta^v$ and $\eta^g$, 
six random numbers are drawn from different random distributions:
$\eta^v_e$ and $\eta^g_e$ are drawn from an exponential distribution with mean 0.05,
$\eta^v_b$ and $\eta^g_b$ are drawn from binominial distributions ${\rm Binom}(1, v^e)$ and ${\rm Binom}(1, g^e)$, 
respectively,
$\eta^v_u$ and $\eta^g_u$ are drawn from an uniform distribution in $[0,1]$.
These random numbers are combined to two noise components $\eta^v$ and $\eta^g$ by
\begin{eqnarray}
\eta^v & = & \eta^v_e + (1 - \eta^v_e) \eta^v_u \eta^v_b v^e\\
\eta^g & = & \eta^g_e + (1 - \eta^g_e) \eta^g_u \eta^g_b g^e\,.
\end{eqnarray}

The latent variables $h_v$ and $h_g$ are caclulated as
\begin{eqnarray}
\label{Eq:hv}
  h_v(t) & = &
  \begin{cases}
    v^e & \text{for}\quad v^e \le 0.05\\
    \min(5,1.1 h_v(t-1)) & \text{for}\quad v^e > 0.05 \land h_v(t-1) \ge 1.2\\
    0.9 h_v(t-1) + \frac{\eta^v}{3} & \text{else}\\
  \end{cases}\\
\label{Eq:hg}
  h_g(t) & = &
  \begin{cases}
    g^e & \text{for}\quad g^e \le 0.05\\
    \min(5,1.1 h_g(t-1)) & \text{for}\quad g^e > 0.05 \land h_g(t-1) \ge 1.2\\
    0.9 h_g(t-1) + \frac{\eta^g}{3} & \text{else}\,.\\
  \end{cases}
\end{eqnarray}
The sub-dynamic of fatigue results in a value for fatigue $f$, 
which is relevant for the reward function. (see Eq. \ref{eq:reward}).

\section{State definitions}
To give an overview on possible state definitions a small summary is given.

\subsection{The observation vector}
Only a part of the state variables is observable. 
This observation vector is also called observable state, 
but one has to keep in mind, that it does not fulfill the markov property.
The observation vector $\vec{O}(t)$ at time $t$ comprises current values of 
velocity $v(t)$, gain $g(t)$, shift $s(t)$, set point $p(t)$, consumption $c(t)$, 
and fatigue $f(t)$.

\begin{table}[h]
  \begin{center}
    \begin{tabular}{lll}
      text name   & symbol & software name\\
      \hline
      set point   & $p(t)$ & {\tt SetPoint}\\
      velocity    & $v(t)$ & {\tt Velocity}\\
      gain        & $g(t)$ & {\tt Gain}\\
      shift       & $s(t)$ & {\tt Shift}\\
      consumption & $c(t)$ & {\tt Consumption}\\
      fatigue     & $f(t)$ & {\tt Fatigue}\\
    \end{tabular}
  \end{center}
  \caption{The observation vector.}
\end{table}

\subsection{The preferred minimal markovian state}
The preferred minimal markovian state fulfills the markov property with the 
minimum number of variables.
It comprises 20 values. 
These are the observation vector (velocity $v(t)$, gain $g(t)$, shift $s(t)$, set point $p(t)$, 
consumption $c(t)$, and fatigue $f(t)$)
plus some latent variables of the sub-dynamics.
The sub-dynamics of operational cost adds a list of previous 
operational costs, $o(t-i)$ with $i\in {1,\cdots,9}$.
Note that the current operational cost $o(t)$ is not part of this state definition.
It would be redundant, as it can be calculated by $v(t)$, gain $g(t)$, and set point $p$.
The sub-dynamics of mis-calibration needs 3 additional latent variables, 
$m_1$, $m_2$, and $m_3$, (Sec. \ref{sec:Dynamics-of-mis-calibration}).
The sub-dynamics of fatigue adds 2 additional latent variables $h_v$ and $h_g$,
(Eq. \ref{Eq:hv} and \ref{Eq:hg}).

\begin{table}[h]
  \begin{center}
    \begin{tabular}{lll}
      text name or description   & symbol & software name\\
      \hline
      set point   & $p(t)$ & {\tt SetPoint}\\
      velocity    & $v(t)$ & {\tt Velocity}\\
      gain        & $g(t)$ & {\tt Gain}\\
      shift       & $s(t)$ & {\tt Shift}\\
      consumption & $c(t)$ & {\tt Consumption}\\
      fatigue     & $f(t)$ & {\tt Fatigue}\\
      operational cost at $t-1$ & $o(t-1)$ & {\tt OperationalCost\_1}\\
      operational cost at $t-2$ & $o(t-2)$ & {\tt OperationalCost\_2}\\
      operational cost at $t-3$ & $o(t-3)$ & {\tt OperationalCost\_3}\\
      operational cost at $t-4$ & $o(t-4)$ & {\tt OperationalCost\_4}\\
      operational cost at $t-5$ & $o(t-5)$ & {\tt OperationalCost\_5}\\
      operational cost at $t-6$ & $o(t-6)$ & {\tt OperationalCost\_6}\\
      operational cost at $t-7$ & $o(t-7)$ & {\tt OperationalCost\_7}\\
      operational cost at $t-8$ & $o(t-8)$ & {\tt OperationalCost\_8}\\
      operational cost at $t-9$ & $o(t-9)$ & {\tt OperationalCost\_9}\\
      $1^{\rm st}$ latent variable of mis-calibration & $m_1^l$ & {\tt MisCalibrationDomain}\\
      $2^{\rm nd}$ latent variable of mis-calibration & $m_2^l$ & {\tt MisCalibrationSystemResponse}\\
      $3^{\rm rd}$ latent variable of mis-calibration & $m_3^l$ & {\tt MisCalibrationPhiIdx}\\
      $1^{\rm st}$ latent variable fatigue & $h_v$ & {\tt FatigueLatentV}\\
      $2^{\rm nd}$ latent variable fatigue & $h_g$ & {\tt FatigueLatentG}\\
    \end{tabular}
  \end{center}
  \caption{The preferred minimal markovian state. 
    It fulfills the markov property with the minimum number of variables.}
\end{table}

\subsection{The extended state}
The extended state, also called the internal markovian state,
contains in addition to all the variables of the preferred minimal markovian state
also some variables, which are of useful for data analysis purposes.

\begin{table}[hp]
  \begin{center}
    \begin{tabular}{lll}
      text name or description   & symbol & software name\\
      \hline
      set point   & $p(t)$ & {\tt SetPoint}\\
      velocity    & $v(t)$ & {\tt Velocity}\\
      effective velocity & $v^e$ & {\tt EffectiveVelocity}\\
      gain        & $g(t)$ & {\tt Gain}\\
      effective gain     & $g^e$ & {\tt EffectiveGain}\\
      shift       & $s(t)$ & {\tt Shift}\\
      effective shift & $s^e(t)$ & {\tt EffectiveShift}\\
      $1^{\rm st}$ latent variable of mis-calibration & $m_1^l$ & {\tt MisCalibrationDomain}\\
      $2^{\rm nd}$ latent variable of mis-calibration & $m_2^l$ & {\tt MisCalibrationSystemResponse}\\
      $3^{\rm rd}$ latent variable of mis-calibration & $m_3^l$ & {\tt MisCalibrationPhiIdx}\\
      mis-calibration & $m(t)$ & {\tt MisCalibration}\\
      modified operational cost & $\hat{c}$ &  {\tt NoiseFreeConsumption}\\
      consumption & $c(t)$ & {\tt Consumption}\\
      fatigue     & $f(t)$ & {\tt Fatigue}\\
      current operational cost  & $o(t)$   & {\tt OperationalCost\_0}\\
      operational cost at $t-1$ & $o(t-1)$ & {\tt OperationalCost\_1}\\
      operational cost at $t-2$ & $o(t-2)$ & {\tt OperationalCost\_2}\\
      operational cost at $t-3$ & $o(t-3)$ & {\tt OperationalCost\_3}\\
      operational cost at $t-4$ & $o(t-4)$ & {\tt OperationalCost\_4}\\
      operational cost at $t-5$ & $o(t-5)$ & {\tt OperationalCost\_5}\\
      operational cost at $t-6$ & $o(t-6)$ & {\tt OperationalCost\_6}\\
      operational cost at $t-7$ & $o(t-7)$ & {\tt OperationalCost\_7}\\
      operational cost at $t-8$ & $o(t-8)$ & {\tt OperationalCost\_8}\\
      operational cost at $t-9$ & $o(t-9)$ & {\tt OperationalCost\_9}\\
      convoluted operational cost & $o^c(t)$ & {\tt OperationalCostConv}\\
      modified operational cost & $\hat{c}(t)$ & {\tt ModifiedOperationalCost}\\
      $1^{\rm st}$ latent variable fatigue & $h_v$ & {\tt FatigueLatentV}\\
      $2^{\rm nd}$ latent variable fatigue & $h_g$ & {\tt FatigueLatentG}\\
    \end{tabular}
  \end{center}
  \caption{The extended state.}
\end{table}

\section{Experimental setup}
\label{sec:Experimental_setup}
To test the algorithms in an initial batch mode, off-policy setting, data is generated by the maximum entropy policy, 
\begin{equation}
\frac{dP(a|s)}{da} = \rm{const}
\end{equation}
The benchmark is initialized for ten different set points $p\in{10,20,\cdots,90,100}$ with the latent variables 
in their default values and the three steerings at 50 each. Then for each set point value the maximum entropy policy
is applied on the benchmark for 1000 time steps, resulting in 10.000 data points.
This data can then be used to train system identification models and/or policies.
The goal is to build a policy $\pi$ that maximizes the average reward on the same setting, 
where instead of the maximum entropy policy, the policy $\pi$ is applied.

\section{Interfaces}
The main interfaces are defined in {\tt com.siemens.rl.interfaces}. They are
DataVector, Environment, and ExternalDriver.

\subsection{Interface DataVector}
This interface lists all necessary methods to implement a data vector, which might be a state- or action-vector.

\begin{arbp}
Modifier and Type & Method and Description\\
{\tt DataVector}        &  {\tt clone()}\\
                        & Returns a copy of the data vector.\\
\\
{\tt List $<$String$>$} & {\tt getKeys()}\\
                        & Returns a list containing the data-vector dimension names.\\
\\
{\tt Double}            & {\tt getValue(String\_key)}\\
                        & Returns the value for a given data-vector dimension.\\
\\
{\tt double[]}          & {\tt getValuesArray()}\\
                        & Returns a double[] array containing the values.\\
\\
{\tt void}              & {\tt setValue(String\_key, double\_value)}\\
                        & Sets the current value of a given data-vector dimension.\\
\end{arbp}

\subsection{Interface Environment}

This interface describes all relevant methods for implementing the dynamics of an environment.

\begin{arbp}
Modifier and Type & Method and Description\\
{\tt void}              &  {\tt addExternalDriver(ExternalDriver\_extDriver)}\\
                        & This function adds an external driver to the environment, 
                          which affect/filter state variables during the call of step().\\
\\
{\tt DataVector}        & {\tt getInternalMarkovState()}\\
                        & Returns the internal Markovian state.\\
\\
{\tt double}            & {\tt getReward()}\\
                        & Returns the reward.\\
\\
{\tt DataVector}        & {\tt getState()}\\
                        & Returns the observable state.\\
\\
{\tt void}              & {\tt reset()}\\
                        & Function for resetting the environment.\\
\\
{\tt double}            & {\tt step(DataVector\_action)}\\
                        & Performs an action within the environment and returns the reward.\\
\end{arbp}

\subsection{Interface ExternalDriver}
Abstract interface for attaching external drivers to the Environment, 
that affect/filter certain state dimensions (e.g. such as set point).

\begin{arbp}
Modifier and Type & Method and Description\\
{\tt void}              &  {\tt filter(DataVector\_state)}\\
                        & Applies "in-place" the external drivers to the given data vector.\\
\\
{\tt DataVector}        & {\tt getState()}\\
                        & Returns the current configuration.\\
\\
{\tt void}              & {\tt setConfiguration()}\\
                        & Sets the external driver configuration from within the given data vector.\\
\\
{\tt void}              & {\tt setSeed(long\_seed)}\\
                        & Sets the random seed.\\
\end{arbp}

\section{Example usage}

\subsection{Class ExampleMain}
An example is implemented in {\tt com.siemens.industrialsim.ExampleMain}.

\begin{arbp}
Modifier and Type & Method and Description\\
{\tt static void}       & {\tt main(String[]\_args)}\\
                        & Run example simulation with random actions.\\
\end{arbp}

\section{First results}
First results with 10.000 data vectors as described in section \ref{sec:Experimental_setup}
indicate, that the reward (named {\tt RewardTotal} can be estimated from current and past values of 
velocity, gain, shift, and set point as inputs by a recurrent neural network 
with a mean relative absolute deviation (MRABD) of approximately 10\%.
Consumption $c$ can be estimated with a MRABD of approximately 3.6\%,
and fatigue $f$ with a MRABD of approximately 24\% (for $f>1$).

The average reward of the maximum entropy policy in the setting described 
in section \ref{sec:Experimental_setup} is -$290.8 \pm 0.6$ with a standard deviation of $20$.


First results with the policy gradient neural rewards regression (PGNRR) \cite{schneegass2007icann} 
and with an extension to continues actions of the neural fitted Q-iteration (NFQ) \cite{riedmiller2005nfqi} 
lead to average rewards of roughly -270.


\newpage
\bibliographystyle{unsrt}
\bibliography{ib}

\begin{thebibliography}{1}

\bibitem{sutton_and_barto1998rl}
Richard~S. Sutton and Andrew~G. Barto.
\newblock {\em Reinforcement Learning: An Introduction}.
\newblock MIT Press, Cambridge, 1998.

\bibitem{Thrun96a}
S.~Thrun.
\newblock Is learning the $n$-th thing any easier than learning the first?
\newblock In D.~Touretzky and M~Mozer, editors, {\em Advances in Neural
  Information Processing Systems (NIPS) 8}, pages 640--646, Cambridge, MA,
  1996. MIT Press.

\bibitem{PanYang2010}
S.~J. Pan and Q.~Yang.
\newblock A survey on transfer learning.
\newblock {\em IEEE Transactions on Knowledge and Data Engineering},
  22(10):1345--1359, Oct 2010.

\bibitem{schneegass2007icann}
Daniel Schneega{\ss}, Steffen Udluft, and Thomas Martinetz.
\newblock Improving optimality of neural rewards regression for data-efficient
  batch near-optimal policy identification.
\newblock In Joaquim~Marques de~S{\'a}, Lu{\'i}s~A. Alexandre, W{\l}odzis{\l}aw
  Duch, and Danilo Mandic, editors, {\em Artificial Neural Networks -- ICANN
  2007: 17th International Conference, Porto, Portugal, September 9-13, 2007,
  Proceedings, Part I}, pages 109--118, Berlin, Heidelberg, 2007. Springer
  Berlin Heidelberg.

\bibitem{riedmiller2005nfqi}
Martin Riedmiller.
\newblock Neural fitted {Q}-iteration - first experiences with a data efficient
  neural reinforcement learning method.
\newblock In {\em Proceedings of the 16th European Conference on Machine
  Learning}, pages 317--328, 2005.

\end{thebibliography}

\end{document}